\documentclass[a4paper,twoside]{article}

\usepackage{epsfig}
\usepackage{subcaption}
\usepackage{calc}
\usepackage{amssymb}
\usepackage{amstext}
\usepackage{amsmath}
\usepackage{amsthm}
\usepackage{multicol}
\usepackage{pslatex}
\usepackage{apalike}
\usepackage[bottom]{footmisc}
\usepackage{SCITEPRESS}     
\usepackage{multirow}
\usepackage[table,xcdraw]{xcolor}
\usepackage{subcaption}
\usepackage[export]{adjustbox}
\begin{document}
\title{FPCD: An Open Aerial VHR Dataset for Farm Pond Change Detection}
\author{\authorname{Chintan Tundia\sup{1}, Rajiv Kumar\sup{1}, Om Damani \sup{1} and G. Sivakumar \sup{1}}
\affiliation{\sup{1}Indian Institute of Technology Bombay, Mumbai, INDIA}
\email{\{chintan, rajiv, damani\}@cse.iitb.ac.in, siva@iitb.ac.in}
}

\abstract{Change detection for aerial imagery involves locating and identifying changes associated with the areas of interest between co-registered bi-temporal or multi-temporal images of a geographical location. 
Farm ponds are man-made structures belonging to the category of minor irrigation structures used to collect surface run-off water for future irrigation purposes. Detection of farm ponds from aerial imagery and their evolution over time helps in land surveying to
analyze the agricultural shifts, policy implementation, seasonal effects and climate changes. In this paper, we introduce a publicly available object detection and instance segmentation (OD/IS) dataset for localizing farm ponds from aerial imagery. We also collected and annotated the bi-temporal data over a time-span of 14 years across 17 villages, resulting in a binary change detection dataset called \textbf{F}arm \textbf{P}ond \textbf{C}hange \textbf{D}etection Dataset (\textbf{FPCD}). 
 We have benchmarked and analyzed the performance of various object detection and instance segmentation methods on our OD/IS dataset and the change detection methods over the FPCD dataset. The datasets are publicly accessible at this page: 
\textit{\url{https://huggingface.co/datasets/ctundia/FPCD}}. 
}
\onecolumn \maketitle \normalsize \vfill
\section{\uppercase{Introduction}}
\label{sec:introduction}
Accurate and timely detection of geographical changes on earth's surface gives extensive information about the various activities and phenomena happening on earth. Change detection task helps in analyzing and understanding co-registered images for change information. A change instance between two images refers to the semantic level differences in the appearance between the two images in association to the regions of interest captured at different points in time. On geographical images, it helps to keep track of the changes to analyze evolution of land geography or land objects and to mitigate hazards at local and global scales. The availability of high-resolution aerial imagery has enabled land use and land cover monitoring to detect objects such as wells, farm ponds, check dams, etc. at instance level. 
\footnote{Authors \textit{a,b} made equal contribution }
\par
Change detection can be bi-temporal when two points in time are compared or multi-temporal when multiple points in time are compared. When multi-temporal data is captured by satellites, drones or aerial vehicles, they are constrained by spatial, spectral and temporal elements, in addition to atmospheric conditions, resolution, etc. The cloud cover, shadows and seasonal changes become visible in many aerial images that affect the overall appearance of the images resulting in the co-registered images to appear from different domains. Moreover, acquiring and annotating images from satellite images to build change detection datasets is a costly process that involves many underlying tasks. The absence of paired images of the same location captured at different times makes it difficult to obtain a useful change detection dataset. Also when paired bi-temporal images are available, there might not be changes present, or the bi-temporal images are captured in very short intervals.
\begin{figure}[ht]
\centering
\setkeys{Gin}{width=0.24\linewidth}
\includegraphics{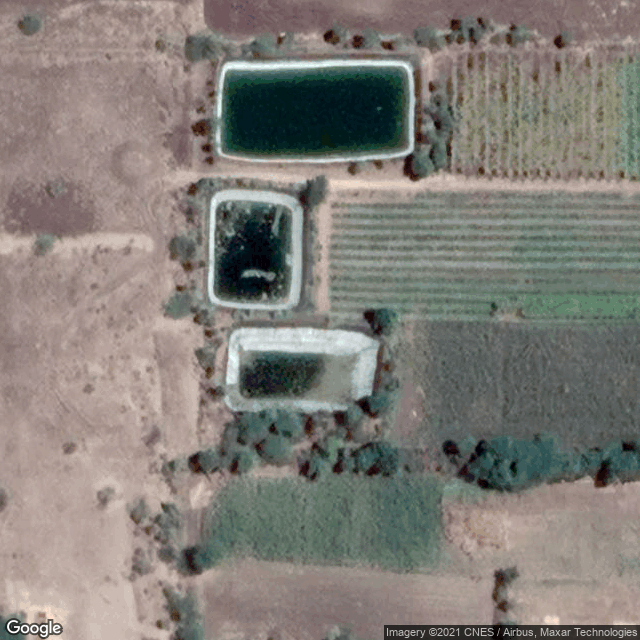}\,%
\includegraphics{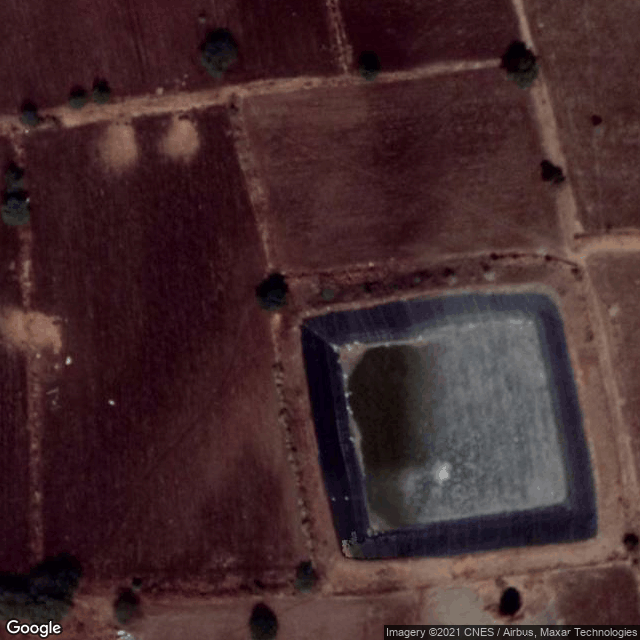}\,%
\includegraphics{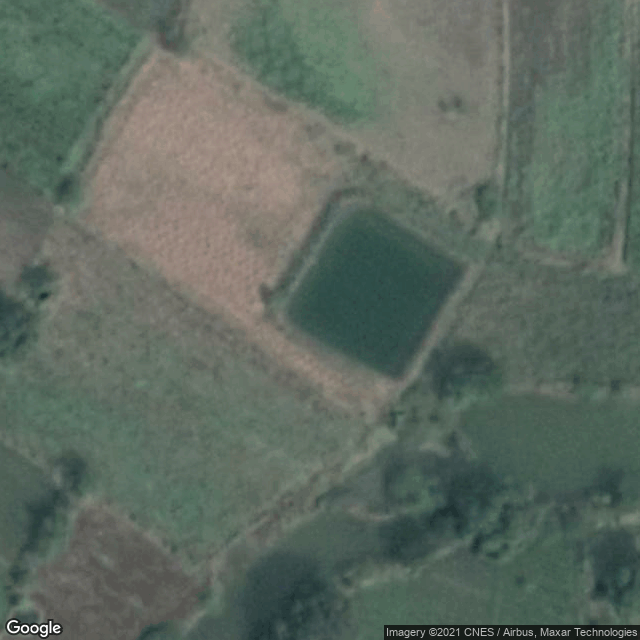}\,%
\includegraphics{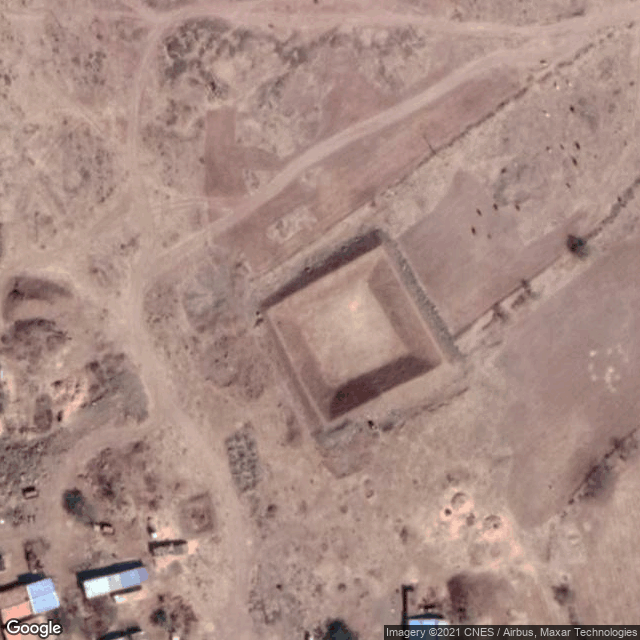}
\caption{Different categories of farm ponds. From left to right: Wet farm pond (lined), Dry farm pond (lined), Wet farm pond (unlined) and Dry farm pond (unlined).}
\label{fig:object_classes}
\end{figure} 
\par
Farm ponds have become popular as private irrigation sources in developing countries like India over the last two decades. A farm pond is an artificial dug out structure having an inlet and outlet for collecting the surface runoff water flowing from the farm area \cite{gistam20}. They are used to collect and store rainwater so as to provide irrigation to crops during periods of water scarcity. It is one of the many minor irrigation structures \cite{visapp22} with a cultivable command area of up to 2000 hectares. Farm ponds can be classified into two categories based on the presence or absence of water in it: wet farm ponds and dry farm ponds. Also, farm ponds can be either lined or unlined depending on the use of plastic lining to prevent water seepage into groundwater. Overall, farm ponds can be classified into 4 sub-categories (See Fig \ref{fig:object_classes}): lined wet farm pond, unlined wet farm pond, lined dry farm pond and unlined dry farm pond based on their structure and the presence of water.
\par
Though efforts have been put up to promote farm ponds, serious concerns have been raised over their implementation and their usage. The purpose of building farm ponds has long drifted from their original objective of storing rainwater for protective irrigation to being used as storage tanks for pumped-out groundwater exposing the underground water to evaporation losses. In the due time, farm ponds have accelerated the rate of groundwater exploitation to many folds \cite{PRASAD2022107385}. 
The objective of our work is to develop change detection models to detect and visualize changes associated with farm ponds and to compute percentage differences in the presence/absence of wet/dry farm ponds. This can help in making policy decisions and in analyzing impacts of farm ponds on aspects like shifting of agricultural practice, policies being implemented, seasonal effects and changes, etc.

\subsection{Contributions}
Our contributions in this paper are:
\begin{enumerate}
\item \textbf{Farm pond change detection dataset (FPCD)}: A publicly available dataset for change detection tasks on farm pond categories for different purposes and stakeholders. 
\item A small-scale public dataset for object detection and instance segmentation of four farm pond categories.
\end{enumerate}

The paper is organized as follows: section~\ref{subsec:related_work} covers the related work,
section~\ref{sec:dataset} covers the details of the proposed dataset,
section \ref{sec:experiments} covers the experiments, section \ref{sec:results} covers the results and observation, and finally conclusion in section \ref{sec:conclusion}.
\begin{table}[h]
\caption{Comparison of change detection datasets.}
\label{tab:datasets}
\centering
\resizebox{\columnwidth}{!}{%
\begin{tabular}{|c|c|c|c|c|c|c|}
\hline
\textbf{Dataset} &
  \textbf{\begin{tabular}[c]{@{}c@{}}Image \\ Pairs\end{tabular}} &
  \textbf{\begin{tabular}[c]{@{}c@{}}Res. \\ (m)\end{tabular}} &
  \textbf{\begin{tabular}[c]{@{}c@{}}Image \\ Size\end{tabular}} &
  \textbf{Bands} &
  \textbf{Mask Type} &
  \textbf{\begin{tabular}[c]{@{}c@{}}Change \\ Inst.\end{tabular}} \\ \hline
  \begin{tabular}[c]{@{}c@{}}CLCD \\ \cite{9780164}\end{tabular}
   &
  600 &
  0.5 to 2 &
  512 x 512 &
  RGB &
  Binary &
   - \\ \hline
   \begin{tabular}[c]{@{}c@{}}S2Looking \\ \cite{rs13245094}\end{tabular}
  &
  5000 &
  \begin{tabular}[c]{@{}c@{}}0.5 \\ to \\ 0.8\end{tabular} &
  1024 x 1024 &
  RGB &
  Binary &
  65,920 \\ \hline
  \begin{tabular}[c]{@{}c@{}}SYSU-CD \\ \cite{9467555}\end{tabular}
  &
  20000 &
  0.5 &
  256 x 256 &
  RGB &
  Binary &
   - \\ \hline
   \begin{tabular}[c]{@{}c@{}}DSIFN \\ \cite{ZHANG2020183}\end{tabular}
  &
  3988 &
  - &
  512 x 512 &
  RGB &
  Binary &
   - \\ \hline
   \begin{tabular}[c]{@{}c@{}}LEVIR-CD  \\ \cite{rs12101662} \end{tabular}
 &
  637 &
  0.5 &
  1024 x 1024 &
  RGB &
  Binary &
  31,333 \\ \hline
\begin{tabular}[c]{@{}c@{}}WHU \\ Building \\ CD \cite{8444434}\end{tabular} &
  1 &
  0.075 &
  \begin{tabular}[c]{@{}c@{}}32207 \\ x \\ 15354\end{tabular} &
  RGB &
  Binary &
  2297 \\ \hline
  \begin{tabular}[c]{@{}c@{}}SZTAKI \\ \cite{6050150} \end{tabular}
  &
  13 &
  1.5 &
  952 x 640 &
  RGB &
  Binary &
  382 \\ \hline
\textbf{FPCD} &
  \textbf{694} &
  \textbf{0.156} &
  \textbf{1024 x 768} &
  \textbf{RGB} &
  \textbf{\begin{tabular}[c]{@{}c@{}}  Binary \end{tabular}} &
  \textbf{616} \\ \hline
\end{tabular}%
}
\end{table}
\subsection{Problem Formulation}
Generally, change detection tasks involve an input set of multi-temporal images and the corresponding ground truth mask, with most change detection datasets having bi-temporal images mapped to binary mask labels. A general assumption in most change detection tasks is that there is a pixel-to-pixel correspondence between the two images and these correspondences are registered to the same point on a geographical area. Based on the correspondences, each pixel in the change mask can be assigned a label indicating whether there is a change or not. In other words, pixels belonging to the change mask are assigned a change label if the corresponding area of interest has geographical changes with respect to each other and those are not assigned a change label when there are no changes. In a binary change detection setting, the paired input images are $T_{0}\in R^{CxHxW}$ and $T_{1}\in R^{CxHxW}$ with a size of CxHxW, where H and W are the spatial dimensions and C=3 is the input image channel dimensions. The ground truth mask can be represented as pixel based mask label, $M \in R^{CxHxW}$ for the bi-temporal input images of the change detection task. 

\subsection{Related Work}
\label{subsec:related_work}
\subsubsection{Datasets}
Change detection datasets generally use RGB images \cite{rs12101662} or hyperspectral images \cite{8518015} with some datasets \cite{Van_Etten_2021_CVPR} having change instances up to 11 million. The image resolution of CD datasets range from a few centimeters \cite{8444434}, \cite{rs13183750}, \cite{DBLP:journals/corr/abs-2011-03247} to 10 meters \cite{Van_Etten_2021_CVPR}, \cite{8518015}. Some of these datasets have image pairs ranging from a few hundred \cite{9780164} to tens of thousands  \cite{9467555} and some have very few input image pairs of very high resolution. While most datasets have fixed image sizes. Most of the CD datasets have modest image sizes except for the exception of a few \cite{7817860}, \cite{8444434}. We summarize the different aspects of various change detection datasets along with their details in Table \ref{tab:datasets}.


\subsubsection{Change Detection Techniques}
Deep learning based change detection methods can be classified into feature-based, patch-based and image based deep learning change detection. 
Earlier CNN architectures used siamese with triplet loss \cite{8488487} and weighted contrastive loss \cite{8022932} to learn discriminative features between change and no-change images, and used Euclidean distances between the images features to generate the difference images. Recent developments in deep learning have led to the usage of attention mechanisms and feature fusion at various scales improving feature extraction capabilities. We compare and list some of the existing encoder-decoder based models used for change detection below. 
\par
Deeplabv3+ \cite{chen2018encoder} uses spatial pyramid pooling module to encode multi-scale context at multiple effective fields-of-view and encode-decoder structure to capture sharper object boundaries. It improves upon DeepLabv3 \cite{Chen2017RethinkingAC} by applying depthwise separable convolution to both Atrous Spatial Pyramid Pooling and decoder modules. Pyramid scene parsing network (PSPNet)\cite{zhao2017pyramid} along with pyramid pooling module applies different-region-based context aggregation to produce global prior representation for pixel-level prediction tasks. Unified Perceptual Parsing Net (UPerNet)\cite{xiao2018unified} is a multi-task framework that can recognize visual concepts from a given image using a training strategy developed to learn from heterogeneous image annotations. Multi-scale Attention Net (MA-Net) \cite{9201310} uses multi-scale feature fusion to improve the segmentation performance by introducing self-attention mechanism to integrate local features with their global dependencies. It uses Position-wise Attention Block (PAB) to model the feature inter-dependencies in spatial dimensions and Multi-scale Fusion Attention Block (MFAB) to capture the channel dependencies between any feature map by multi-scale semantic feature fusion. 
\par
LinkNet \cite{Chaurasia2017LinkNetEE} proposes a novel deep neural network architecture that uses only 11.5 million parameters learning without any significant increase in number of parameters. Pyramid Attention Network (PAN) \cite{li2018pyramid} uses Feature Pyramid Attention module and Global Attention Upsample module to combine attention mechanism and spatial pyramid to extract precise dense features for pixel labeling. Unet++ \cite{zhou2018unet++} improves on Unet \cite{ronneberger2015u} with new skip pathways and is based on the idea that optimizer can learning easily with semantically similar feature maps by reducing the gap between the feature maps of the encoder and decoder sub-networks. Bitemporal image transformer (BiT) \cite{chen2021a} models contexts within the spatial-temporal domain in a deep feature differencing-based CD framework. The bi-temporal images are encoded as tokens and a transformer encoder models the contexts in the compact token-based space-time. A transformer decoder then refines the original features from the learned context-rich tokens back to the pixel-space. 
\subsubsection{Object Detection and Instance Segmentation Techniques}
Object detection is a computer vision task that localizes and classifies objects of interest in an image. Instance segmentation on the other hand provides a detailed inference for every single pixel in input image.
Since the advent of deep learning\cite{rajiv_VISAPP21}, object detection has vastly benefited from sophisticated architectures and image representations.
Over the years, deep learning based detectors have evolved into one-stage and two-stage detectors. In one-stage detection, the input image is divided into regions simultaneously with the probabilistic prediction of objects, while in two-stage object detection the object proposals are classified in the second stage from a sparse set of candidate object proposals generated in the first stage. A few of the well-known two-stage detectors include the FasterRCNN \cite{ren2015faster}, GridRCNN \cite{lu2019grid}, while the YOLOv3 \cite{redmon2018yolov3}, Generalized Focal Loss \cite{li2020gfl}, Gradient Harmonized SSD \cite{li2019gradient} are one-stage detectors. Detecting objects from aerial imagery is affected by adversities like viewpoint variation, illumination, occlusion, etc. and becomes difficult due to objects being small, sparse and non-uniform in high-resolution aerial images.


\begin{table}[]
\centering
\caption{Temporal object instances and their respective change classes. Row 1 corresponds to T0 images, row 2 corresponds to T1 images, row 3 corresponds to change mask. Columns - (a,b) No Change, (c) Farm pond constructed, (d) Farm pond demolished, (e) Farm pond dried and (f) Farm pond wetted.} 
\resizebox{0.95\columnwidth}{!}{%
\begin{tabular}{ccccccc}

\includegraphics[width=.3\columnwidth,valign=m]{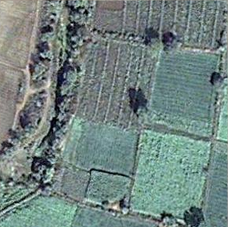} & \includegraphics[width=.3\linewidth,valign=m]{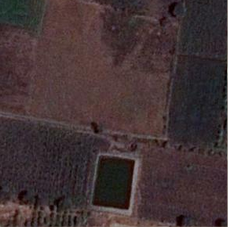} & \includegraphics[width=.3\linewidth,valign=m]{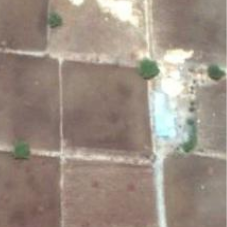} &
\includegraphics[width=.3\linewidth,valign=m]{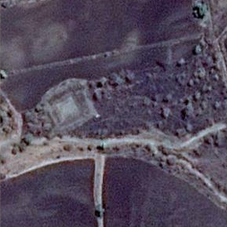} &
\includegraphics[width=.3\linewidth,valign=m]{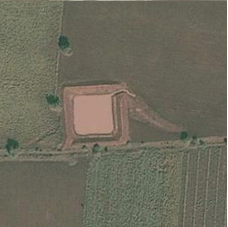}&
\includegraphics[width=.3\linewidth,valign=m]{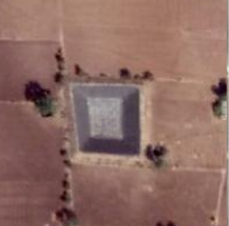} 
\\[0.9cm] 
\includegraphics[width=.3\linewidth,valign=m]{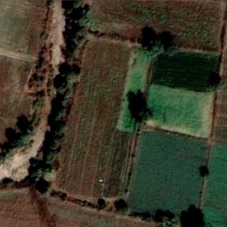} & \includegraphics[width=.3\linewidth,valign=m]{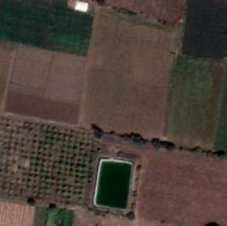} & \includegraphics[width=.3\linewidth,valign=m]{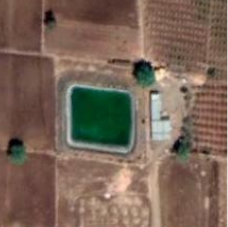} &
\includegraphics[width=.3\linewidth,valign=m]{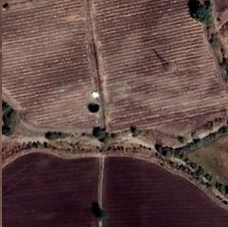} &
\includegraphics[width=.3\linewidth,valign=m]{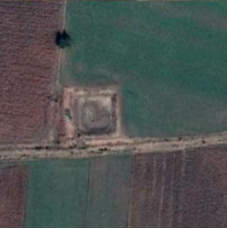}&
\includegraphics[width=.3\linewidth,valign=m]{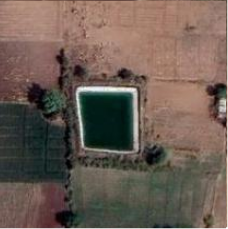} 
\\[0.9cm] 
\includegraphics[width=.3\linewidth,valign=m]{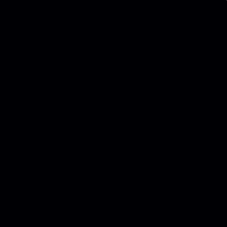} & \includegraphics[width=.3\linewidth,valign=m]{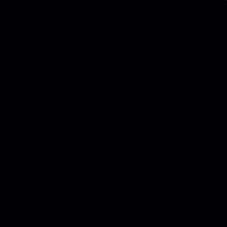} & \includegraphics[width=.3\linewidth,valign=m]{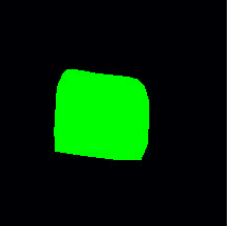} &
\includegraphics[width=.3\linewidth,valign=m]{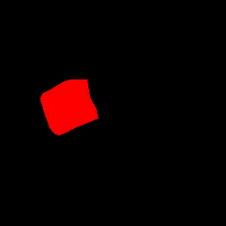} &
\includegraphics[width=.3\linewidth,valign=m]{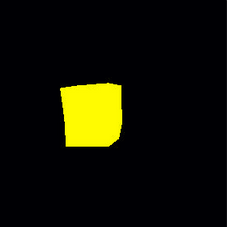}&
\includegraphics[width=.3\linewidth,valign=m]{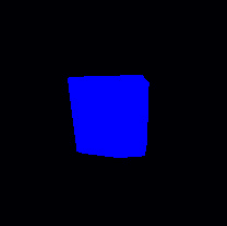} 
\\[0.9cm]
\textbf{a} & \textbf{b} & \textbf{c} & \textbf{d} & \textbf{e} & \textbf{f}
\end{tabular}
}
\label{tab:classchange}
\end{table}

\section{Farm pond Change Detection (FPCD) Dataset}
\label{sec:dataset}
\noindent
FPCD is a novel publicly available change detection dataset that focuses on changes associated with irrigation structures from India. 
The details of the multi-temporal images including the district, village name and the time-interval over which the images were collected are summarized in Table \ref{tab:datasetStats}.

\begin{table}[h]
\caption{Class distribution of change objects.}
\label{tab:datasetClassDist}
\centering
\resizebox{0.4\columnwidth}{!}{%
\begin{tabular}{|c|c|}
\hline
\textbf{\begin{tabular}[c]{@{}c@{}}Change\\ Class\end{tabular}} &
  \textbf{\begin{tabular}[c]{@{}c@{}}No. of \\ Instances\end{tabular}} \\ \hline
Farm pond constructed & 431  \\ \hline
Farm pond demolished  & 39  \\ \hline
Farm pond dried & 47  \\ \hline
Farm pond wetted      & 99  \\ \hline
\end{tabular}%
}
\end{table}
\begin{table}[h]
\caption{Class distribution of object instances in OD/IS dataset.}
\label{tab:datasetObjDist}
\centering
\resizebox{0.4\columnwidth}{!}{%
\begin{tabular}{|c|c|}
\hline
\textbf{\begin{tabular}[c]{@{}c@{}}Annotation \\ Class\end{tabular}} & \textbf{\begin{tabular}[c]{@{}c@{}}No. of \\ Instances\end{tabular}} \\ \hline
Wet farm pond (lined)  & 287 \\ \hline
Wet farm pond (unlined) & 293 \\ \hline
Dry farm pond (lined)& 90  \\ \hline
Dry farm pond (unlined)& 668 \\ \hline
\end{tabular}%
}
\end{table}

\begin{table}[h]
\caption{Villages and timestamps for FPCD dataset.}
\label{tab:datasetStats}
\centering
\resizebox{0.75\columnwidth}{!}{
\begin{tabular}{|c|c|c|c|}
\hline
\textbf{District} &
  \textbf{Village Name} &
  \textbf{Timestamp} &
  \textbf{\begin{tabular}[c]{@{}c@{}}\#Image \\ Pairs\end{tabular}} \\ \hline
Nasik      & Hadap Sawargaon & Feb-2019, Feb-2021      & 25 \\ \hline
Akola      & Akhatwada & Mar-2007, Mar-2018      & 24 \\ \hline
Akola      & Ghusar & April-2013, Mar-2018    & 124 \\ \hline
Amravati   & Nardoda  & Nov-2009, Jan-2021      & 39 \\ \hline
Aurangabad & Kumbephal & Feb-2014, Mar-2020      & 62 \\ \hline
Beed & Kumbhephal      & April-2012, Mar-2019    & 37 \\ \hline
Buldhana   & Bhivgaon & Jan-2014, Jan-2019      & 8  \\ \hline
Hingoli    & Gondala  & Feb-2014, Mar-2020      & 46 \\ \hline
Jalgaon    & Pimpalgaon Block  & Oct-2013, Nov-2020      & 17 \\ \hline
Jalna &
  \begin{tabular}[c]{@{}c@{}}Dawargaon/\\ Bhatkheda\end{tabular} &
  April-2012, April-2019 &
  16 \\ \hline
Latur      & Sumthana & Jan-2013, Jan-2021      & 15 \\ \hline
Nanded     & Chainpur & Jan-2014, Mar-2019      & 56 \\ \hline
Osmanabad  & Ambejawalga     & Jan-2014, Feb-2019      & 74 \\ \hline
Parbhani   & Kinhola Block      & May-2010, April-2020    & 62 \\ \hline
Wardha     & Kakaddara & Jan-2013, Feb-2019      & 17 \\ \hline
Washim     & Bhoyata  & April-2013, Feb-2021    & 46 \\ \hline
Yavatmal   & Wadhona Pilki   & April-2014,  April-2019 & 26 \\ \hline
 & & \textbf{Total Images}   & \textbf{694} \\ \hline
\end{tabular}%
}
\end{table}

\subsection{Dataset Collection Technique}
\label{subsec:sec_DataCollTechnique}
The images were collected by following the steps given below:-
\begin{enumerate}
\item \textbf{Area and Timestamp Selection:} The ground truth information about the farm pond locations and the list of villages were collected from different sources like news articles, reports and web-pages along with Jalyukt Shivar data\cite{website:jalyukt-shivar}. The location of the Farm ponds would span across the different districts of Maharashtra, India. The villages were then rigorously filtered and chosen based on the presence or absence of farm ponds, and bi-temporal pairs were decided by visual inspection of various timestamps.
\item \textbf{Grid Formation:} For collecting the images of a fixed size for each village, the grids of latitude and longitude at a fixed zoom level of Google Maps were needed. We selected zoom level as 18 which provides resolution of sub-meter level. Grids of size 1024x768 pixels were created by using the village boundaries provided by Indian Village Boundaries Project \cite{website:vilboundaries} in the geo-location dimensions i.e. latitudes and longitudes. 
\item \textbf{Image Collection:} Google Earth Pro desktop software was used for collecting historical imagery. For each grid cell, the map was set such that it would cover the grid cell boundaries with the view. Then the bi-temporal images were saved by setting the timestamp one at a time using the Google Earth Historical Imagery tool. 
\item \textbf{Object \& Instance Annotation:} Once we have collected all the image pairs for each village, we annotate the objects individually in each image. Depending on the types of farm ponds, there are four main classes as given in Table \ref{tab:datasetObjDist}. We use the annotation tool, LabelMe\cite{russell2008labelme} to annotate and label the above mentioned classes. And further we convert them into CoCo\cite{lin2014microsoft} format required for the object detection and instance segmentation tasks thus forming the Farm pond OD/IS dataset.
\item \textbf{Change Mask Generation:} 
In this step, we take bi-temporal pair of images and corresponding farm pond annotations. Based on the location and farm pond category we generate different types of change masks with change classes as given in Table \ref{tab:classchange}. More details on the type of changes are given in subsection \ref{subsec:dataset_details}.
\end{enumerate}


\subsection{Dataset Details}
\label{subsec:dataset_details}
A total of 694 images of size 1024 x 768 pixels at zoom level 18 were collected from Google Earth images using the technique described in subsection \ref{subsec:sec_DataCollTechnique}. The regions of Maharashtra in India were chosen, since it is largely a groundwater dependent region in western India. The images collected at zoom level 18 are at a very high resolution scale, up to 1 meter. The details of villages collected and their respective timestamps are given in Table \ref{tab:datasetStats}. Most of the villages have timestamps during the months of Jan-April 
and the minimum year difference between bi-temporal images is 2 years and the maximum year difference is 9 years, earliest being 2007 and latest being 2021. The FPCD dataset consists of image pairs, change masks and object annotations of farm ponds as polygons set in COCO \cite{lin2014microsoft} format. 
\par
For farm pond change detection, we identify four change classes i.e Farm pond constructed, Farm pond demolished, Farm pond dried and Farm pond wetted. 
Let's consider the bi-temporal pair as T0-T1, T0 being the image captured with the old timestamp and T1 being the image captured with new timestamp image. 
We identify a binary change as FP constructed, when there is no farm pond in the T0 image, but a farm pond is observed at the same location in the T1 image. Likewise, we identify a binary change as FP demolished, when a farm pond existed at any location in the T0 image and there is no farm pond at the same location or replaced by different terrain in the T1 image. We identify a binary change as farm pond dried when there existed a wet farm pond in the T0 image and the farm pond at the same location in T1 image becomes dry due to absence of water or presence of low water level.  We identify a binary change as farm pond wetted, when there existed a dry farm pond in the T0 image and the farm pond at the same location in T1 image is filled with water or the water reaches the surface level. 
\par
Some farmers may pump groundwater and fill the farm ponds instead of using surface-water runoff due to rain leading to water loss by further evaporation. This leads to depletion of groundwater and the practice of farm pond based irrigation becoming unsustainable \cite{PRASAD2022107385}. Thus grouping farm pond constructed and farm pond demolished class helps in identifying the increase/decrease in farm ponds. This helps stakeholders like researchers and policy makers to monitor the impacts due to increase/decrease in farm ponds like change in agricultural patterns, impact on ground water, etc. We classify this task as Task-1.
\par
In certain dry and semi-arid regions, due to uncertain climatic conditions, yearly and seasonal rainfall is erratic. 
Agricultural officers and researchers often correlate climatic conditions and ground water levels with such changes to infer further observations eventually leading to necessary interventions needed for agricultural sustainability. Thus we group Farm pond dried and Farm pond wetted into Task-2. We also combined the above grouped changes so as to also provide overall statistics. We call this as Task-3. We further explain experimental details of Task-1, Task-2 and Task-3 in Section \ref{sec:experiments}. The details of change class distribution are given in the Table \ref{tab:datasetClassDist}. 

\begin{table}[h]
\centering
\caption{
Distribution of positives and negative image pairs of various change detection tasks.
}
\label{tab:cd_train_test_split}
\resizebox{0.55\columnwidth}{!}{%
\begin{tabular}{|c|cc|cc|cc|}
\hline
& \multicolumn{2}{c|}{\textbf{Task 1}} & \multicolumn{2}{c|}{\textbf{Task 2}} & \multicolumn{2}{c|}{\textbf{Task 3}} \\ \hline
\textbf{Type} & \multicolumn{1}{c|}{Neg} & Pos & \multicolumn{1}{c|}{Neg} & Pos & \multicolumn{1}{c|}{Neg} & Pos \\ \hline
\textbf{Train} & \multicolumn{1}{c|}{363}    & 237    & \multicolumn{1}{c|}{534}     & 66    & \multicolumn{1}{c|}{329}    & 271    \\ \hline
\textbf{Validation}   & \multicolumn{1}{c|}{52}     & 42     & \multicolumn{1}{c|}{83}      & 11    & \multicolumn{1}{c|}{45}     & 49     \\ \hline
\end{tabular}%
}
\end{table}



\begin{table}[h]
\centering
\caption{Binary change detection benchmark on farm pond constructed/demolished task (Task 1).}
\label{tab:results_task1}
\resizebox{0.9\columnwidth}{!}{%
\begin{tabular}{|c|ccc|ccc|ccc|}
\hline
\textbf{} &
  \multicolumn{3}{c|}{\textbf{ResNet - 18}} &
  \multicolumn{3}{c|}{\textbf{ResNet - 50}} &
  \multicolumn{3}{c|}{\textbf{ResNet - 101}} \\ \hline
\textbf{Model} &
  \multicolumn{1}{c|}{\textbf{Precision}} &
  \multicolumn{1}{c|}{\textbf{Recall}} &
  \textbf{F-Score} &
  \multicolumn{1}{c|}{\textbf{Precision}} &
  \multicolumn{1}{c|}{\textbf{Recall}} &
  \textbf{F-Score} &
  \multicolumn{1}{c|}{\textbf{Precision}} &
  \multicolumn{1}{c|}{\textbf{Recall}} &
  \textbf{F-Score} \\ \hline
UNet &
  \multicolumn{1}{c|}{0.8174} &
  \multicolumn{1}{c|}{\textbf{0.8537}} &
  0.7360 &
  \multicolumn{1}{c|}{0.9331} &
  \multicolumn{1}{c|}{0.5533} &
  0.5534 &
  \multicolumn{1}{c|}{0.9051} &
  \multicolumn{1}{c|}{0.7072} &
  0.666 \\ \hline
UNet++ &
  \multicolumn{1}{c|}{0.8467} &
  \multicolumn{1}{c|}{0.8439} &
  0.7668 &
  \multicolumn{1}{c|}{0.8465} &
  \multicolumn{1}{c|}{0.8439} &
  0.7668 &
  \multicolumn{1}{c|}{0.9123} &
  \multicolumn{1}{c|}{0.7969} &
  0.7669 \\ \hline
MANet &
  \multicolumn{1}{c|}{0.8347} &
  \multicolumn{1}{c|}{0.5536} &
  0.5539 &
  \multicolumn{1}{c|}{\textbf{0.9998}} &
  \multicolumn{1}{c|}{0.5531} &
  0.5532 &
  \multicolumn{1}{c|}{\textbf{0.9894}} &
  \multicolumn{1}{c|}{0.5596} &
  0.5539 \\ \hline
LinkNet &
  \multicolumn{1}{c|}{0.8558} &
  \multicolumn{1}{c|}{0.8251} &
  0.7529 &
  \multicolumn{1}{c|}{0.8580} &
  \multicolumn{1}{c|}{0.7991} &
  0.7226 &
  \multicolumn{1}{c|}{0.9029} &
  \multicolumn{1}{c|}{0.7554} &
  0.7233 \\ \hline
FPN &
  \multicolumn{1}{c|}{0.8833} &
  \multicolumn{1}{c|}{0.8106} &
  0.7567 &
  \multicolumn{1}{c|}{0.8559} &
  \multicolumn{1}{c|}{\textbf{0.8446}} &
  0.7788 &
  \multicolumn{1}{c|}{0.8582} &
  \multicolumn{1}{c|}{\textbf{0.8764}} &
  0.8003 \\ \hline
{\color[HTML]{24292F} PSPNet} &
  \multicolumn{1}{c|}{\textbf{0.9763}} &
  \multicolumn{1}{c|}{0.5855} &
  0.5840 &
  \multicolumn{1}{c|}{0.9763} &
  \multicolumn{1}{c|}{0.5855} &
  0.5840 &
  \multicolumn{1}{c|}{0.9496} &
  \multicolumn{1}{c|}{0.6178} &
  0.6172 \\ \hline
{\color[HTML]{24292F} PAN} &
  \multicolumn{1}{c|}{0.8572} &
  \multicolumn{1}{c|}{0.8037} &
  0.7250 &
  \multicolumn{1}{c|}{0.8546} &
  \multicolumn{1}{c|}{0.8383} &
  0.7591 &
  \multicolumn{1}{c|}{0.9176} &
  \multicolumn{1}{c|}{0.7907} &
  0.7685 \\ \hline
{\color[HTML]{24292F} DeepLabV3} &
  \multicolumn{1}{c|}{0.8474} &
  \multicolumn{1}{c|}{0.8428} &
  0.7776 &
  \multicolumn{1}{c|}{0.9138} &
  \multicolumn{1}{c|}{0.7917} &
  0.7720 &
  \multicolumn{1}{c|}{0.8891} &
  \multicolumn{1}{c|}{0.8188} &
  0.7894 \\ \hline
{\color[HTML]{24292F} DeepLabV3+} &
  \multicolumn{1}{c|}{0.8998} &
  \multicolumn{1}{c|}{0.8085} &
  0.7870 &
  \multicolumn{1}{c|}{0.8953} &
  \multicolumn{1}{c|}{0.8151} &
  0.7717 &
  \multicolumn{1}{c|}{0.8563} &
  \multicolumn{1}{c|}{0.8348} &
  0.7771 \\ \hline
{\color[HTML]{24292F} UPerNet} &
  \multicolumn{1}{c|}{0.8843} &
  \multicolumn{1}{c|}{0.8148} &
  0.7783 &
  \multicolumn{1}{c|}{0.9055} &
  \multicolumn{1}{c|}{0.8209} &
  \textbf{0.7855} &
  \multicolumn{1}{c|}{0.9131} &
  \multicolumn{1}{c|}{0.8389} &
  \textbf{0.8067} \\ \hline
{\color[HTML]{24292F} BiT} &
  \multicolumn{1}{c|}{0.9252} &
  \multicolumn{1}{c|}{0.8283} &
  \textbf{0.8704} &
  \multicolumn{1}{c|}{-} &
  \multicolumn{1}{c|}{-} &
   -&
  \multicolumn{1}{c|}{-} &
  \multicolumn{1}{c|}{-} &
   -\\ \hline
\end{tabular}%
}
\end{table}

\begin{table}[h]
\centering
\caption{Binary change detection benchmark on farm pond dried/wetted task (Task 2).}
\label{tab:results_task2}
\resizebox{0.9\columnwidth}{!}{%
\begin{tabular}{|c|ccc|ccc|ccc|}
\hline
\textbf{} &
  \multicolumn{3}{c|}{\textbf{ResNet - 18}} &
  \multicolumn{3}{c|}{\textbf{ResNet - 50}} &
  \multicolumn{3}{c|}{\textbf{ResNet - 101}} \\ \hline
\textbf{Model} &
  \multicolumn{1}{c|}{\textbf{Precision}} &
  \multicolumn{1}{c|}{\textbf{Recall}} &
  \textbf{F-Score} &
  \multicolumn{1}{c|}{\textbf{Precision}} &
  \multicolumn{1}{c|}{\textbf{Recall}} &
  \textbf{F-Score} &
  \multicolumn{1}{c|}{\textbf{Precision}} &
  \multicolumn{1}{c|}{\textbf{Recall}} &
  \textbf{F-Score} \\ \hline
UNet &
  \multicolumn{1}{c|}{\textbf{0.9973}} &
  \multicolumn{1}{c|}{0.9094} &
  \textbf{0.9171} &
  \multicolumn{1}{c|}{0.9893} &
  \multicolumn{1}{c|}{0.9091} &
  0.9082 &
  \multicolumn{1}{c|}{0.9907} &
  \multicolumn{1}{c|}{0.9128} &
  0.9131 \\ \hline
UNet++ &
  \multicolumn{1}{c|}{0.9813} &
  \multicolumn{1}{c|}{\textbf{0.9188}} &
  0.9112 &
  \multicolumn{1}{c|}{0.978} &
  \multicolumn{1}{c|}{0.9181} &
  0.90573 &
  \multicolumn{1}{c|}{0.9785} &
  \multicolumn{1}{c|}{0.9147} &
  0.9065 \\ \hline
MANet &
  \multicolumn{1}{c|}{0.9826} &
  \multicolumn{1}{c|}{0.9075} &
  0.897 &
  \multicolumn{1}{c|}{0.9927} &
  \multicolumn{1}{c|}{0.9023} &
  0.9052 &
  \multicolumn{1}{c|}{0.9927} &
  \multicolumn{1}{c|}{0.8966} &
  0.902 \\ \hline
LinkNet &
  \multicolumn{1}{c|}{0.9935} &
  \multicolumn{1}{c|}{0.8867} &
  0.8893 &
  \multicolumn{1}{c|}{0.994} &
  \multicolumn{1}{c|}{0.8974} &
  0.9014 &
  \multicolumn{1}{c|}{0.9918} &
  \multicolumn{1}{c|}{0.91} &
  0.9119 \\ \hline
FPN &
  \multicolumn{1}{c|}{0.9923} &
  \multicolumn{1}{c|}{0.9057} &
  0.9063 &
  \multicolumn{1}{c|}{0.9936} &
  \multicolumn{1}{c|}{0.9125} &
  \textbf{0.9125} &
  \multicolumn{1}{c|}{\textbf{0.9949}} &
  \multicolumn{1}{c|}{0.9125} &
  \textbf{0.9133} \\ \hline
{\color[HTML]{24292F} PSPNet} &
  \multicolumn{1}{c|}{0.9939} &
  \multicolumn{1}{c|}{0.9054} &
  0.907 &
  \multicolumn{1}{c|}{0.9948} &
  \multicolumn{1}{c|}{0.9023} &
  0.9052 &
  \multicolumn{1}{c|}{0.9926} &
  \multicolumn{1}{c|}{0.9054} &
  0.9064 \\ \hline
{\color[HTML]{24292F} PAN} &
  \multicolumn{1}{c|}{0.9907} &
  \multicolumn{1}{c|}{0.9146} &
  0.9116 &
  \multicolumn{1}{c|}{0.9905} &
  \multicolumn{1}{c|}{0.9132} &
  0.9113 &
  \multicolumn{1}{c|}{0.9812} &
  \multicolumn{1}{c|}{0.9135} &
  0.901 \\ \hline
{\color[HTML]{24292F} DeepLabV3} &
  \multicolumn{1}{c|}{0.9814} &
  \multicolumn{1}{c|}{0.9066} &
  0.8971 &
  \multicolumn{1}{c|}{0.9326} &
  \multicolumn{1}{c|}{\textbf{0.9246}} &
  0.8698 &
  \multicolumn{1}{c|}{0.9687} &
  \multicolumn{1}{c|}{0.9096} &
  0.8888 \\ \hline
{\color[HTML]{24292F} DeepLabV3+} &
  \multicolumn{1}{c|}{0.9873} &
  \multicolumn{1}{c|}{0.9081} &
  0.908 &
  \multicolumn{1}{c|}{0.9918} &
  \multicolumn{1}{c|}{0.9126} &
  0.9109 &
  \multicolumn{1}{c|}{0.9767} &
  \multicolumn{1}{c|}{\textbf{0.9246}} &
  0.912 \\ \hline
{\color[HTML]{24292F} UPerNet} &
  \multicolumn{1}{c|}{0.9936} &
  \multicolumn{1}{c|}{0.91367} &
  0.9147 &
  \multicolumn{1}{c|}{\textbf{0.9957}} &
  \multicolumn{1}{c|}{0.9115} &
  0.911 &
  \multicolumn{1}{c|}{0.9929} &
  \multicolumn{1}{c|}{0.9106} &
  0.9096 \\ \hline
  {\color[HTML]{24292F} BiT} &
  \multicolumn{1}{c|}{0.8458} &
  \multicolumn{1}{c|}{0.8254} &
  0.8352 &
  \multicolumn{1}{c|}{-} &
  \multicolumn{1}{c|}{-} &
   -&
  \multicolumn{1}{c|}{-} &
  \multicolumn{1}{c|}{-} &
   -\\ \hline
\end{tabular}
}
\end{table}

\begin{table}[h]
\centering
\caption{
Comparison between results of training with all image pairs and training with only positives images on Task 3. 
}
\label{tab: results_task3}
\resizebox{0.95\columnwidth}{!}{%
\begin{tabular}{|c|cccccc|cccccc|}
\hline
 &
  \multicolumn{6}{c|}{\textbf{With all image pairs}} &
  \multicolumn{6}{c|}{\textbf{With only positive image pairs}} \\ \hline
 &
  \multicolumn{3}{c|}{\textbf{ResNet - 18}} &
  \multicolumn{3}{c|}{\textbf{ResNet - 50}} &
  \multicolumn{3}{c|}{\textbf{ResNet - 18}} &
  \multicolumn{3}{c|}{\textbf{ResNet - 50}} \\ \hline
\textbf{Model} &
  \multicolumn{1}{c|}{\textbf{Precision}} &
  \multicolumn{1}{c|}{\textbf{Recall}} &
  \multicolumn{1}{c|}{\textbf{F-Score}} &
  \multicolumn{1}{c|}{\textbf{Precision}} &
  \multicolumn{1}{c|}{\textbf{Recall}} &
  \textbf{F-Score} &
  \multicolumn{1}{c|}{\textbf{Precision}} &
  \multicolumn{1}{c|}{\textbf{Recall}} &
  \multicolumn{1}{c|}{\textbf{F-Score}} &
  \multicolumn{1}{c|}{\textbf{Precision}} &
  \multicolumn{1}{c|}{\textbf{Recall}} &
  \textbf{F-Score} \\ \hline
UNet &
  \multicolumn{1}{c|}{0.8627} &
  \multicolumn{1}{c|}{0.7778} &
  \multicolumn{1}{c|}{0.7444} &
  \multicolumn{1}{c|}{0.864} &
  \multicolumn{1}{c|}{0.7794} &
  0.7442 &
  \multicolumn{1}{c|}{0.798} &
  \multicolumn{1}{c|}{0.7933} &
  \multicolumn{1}{c|}{0.7049} &
  \multicolumn{1}{c|}{0.8163} &
  \multicolumn{1}{c|}{0.7967} &
  0.7253 \\ \hline
UNet++ &
  \multicolumn{1}{c|}{0.8961} &
  \multicolumn{1}{c|}{0.7795} &
  \multicolumn{1}{c|}{0.7579} &
  \multicolumn{1}{c|}{0.8653} &
  \multicolumn{1}{c|}{0.8081} &
  0.7575 &
  \multicolumn{1}{c|}{0.7884} &
  \multicolumn{1}{c|}{0.8008} &
  \multicolumn{1}{c|}{0.7013} &
  \multicolumn{1}{c|}{0.8441} &
  \multicolumn{1}{c|}{0.8482} &
  \textbf{0.768} \\ \hline
MANet &
  \multicolumn{1}{c|}{0.7868} &
  \multicolumn{1}{c|}{0.821} &
  \multicolumn{1}{c|}{0.741} &
  \multicolumn{1}{c|}{0.8451} &
  \multicolumn{1}{c|}{0.7586} &
  0.7246 &
  \multicolumn{1}{c|}{0.7115} &
  \multicolumn{1}{c|}{0.7925} &
  \multicolumn{1}{c|}{0.6388} &
  \multicolumn{1}{c|}{0.7944} &
  \multicolumn{1}{c|}{0.7054} &
  0.6145 \\ \hline
LinkNet &
  \multicolumn{1}{c|}{0.8592} &
  \multicolumn{1}{c|}{0.7872} &
  \multicolumn{1}{c|}{0.7268} &
  \multicolumn{1}{c|}{0.7868} &
  \multicolumn{1}{c|}{\textbf{0.8604}} &
  0.7502 &
  \multicolumn{1}{c|}{0.7667} &
  \multicolumn{1}{c|}{0.8024} &
  \multicolumn{1}{c|}{0.6871} &
  \multicolumn{1}{c|}{0.8753} &
  \multicolumn{1}{c|}{0.7627} &
  0.7236 \\ \hline
FPN &
  \multicolumn{1}{c|}{0.8775} &
  \multicolumn{1}{c|}{0.7856} &
  \multicolumn{1}{c|}{0.7653} &
  \multicolumn{1}{c|}{0.9192} &
  \multicolumn{1}{c|}{0.7591} &
  0.7696 &
  \multicolumn{1}{c|}{\textbf{0.9067}} &
  \multicolumn{1}{c|}{0.6843} &
  \multicolumn{1}{c|}{0.6909} &
  \multicolumn{1}{c|}{0.8403} &
  \multicolumn{1}{c|}{0.8094} &
  0.7442 \\ \hline
{\color[HTML]{24292F} PSPNet} &
  \multicolumn{1}{c|}{0.8749} &
  \multicolumn{1}{c|}{0.5652} &
  \multicolumn{1}{c|}{0.5519} &
  \multicolumn{1}{c|}{\textbf{0.9571}} &
  \multicolumn{1}{c|}{0.5767} &
  0.5919 &
  \multicolumn{1}{c|}{0.8518} &
  \multicolumn{1}{c|}{0.589} &
  \multicolumn{1}{c|}{0.5564} &
  \multicolumn{1}{c|}{\textbf{0.8783}} &
  \multicolumn{1}{c|}{0.6212} &
  0.601 \\ \hline
{\color[HTML]{24292F} PAN} &
  \multicolumn{1}{c|}{0.8804} &
  \multicolumn{1}{c|}{0.7252} &
  \multicolumn{1}{c|}{0.7048} &
  \multicolumn{1}{c|}{0.7228} &
  \multicolumn{1}{c|}{0.6935} &
  0.5435 &
  \multicolumn{1}{c|}{0.752} &
  \multicolumn{1}{c|}{0.7977} &
  \multicolumn{1}{c|}{0.6575} &
  \multicolumn{1}{c|}{0.6464} &
  \multicolumn{1}{c|}{0.7536} &
  0.5753 \\ \hline
{\color[HTML]{24292F} DeepLabV3} &
  \multicolumn{1}{c|}{\textbf{0.9329}} &
  \multicolumn{1}{c|}{0.7462} &
  \multicolumn{1}{c|}{0.7652} &
  \multicolumn{1}{c|}{0.8775} &
  \multicolumn{1}{c|}{0.8257} &
  \textbf{0.7952} &
  \multicolumn{1}{c|}{0.7656} &
  \multicolumn{1}{c|}{0.8276} &
  \multicolumn{1}{c|}{0.6813} &
  \multicolumn{1}{c|}{0.7744} &
  \multicolumn{1}{c|}{\textbf{0.8611}} &
  0.7169 \\ \hline
{\color[HTML]{24292F} DeepLabV3+} &
  \multicolumn{1}{c|}{0.8232} &
  \multicolumn{1}{c|}{0.7953} &
  \multicolumn{1}{c|}{0.7238} &
  \multicolumn{1}{c|}{0.9522} &
  \multicolumn{1}{c|}{0.7118} &
  0.7519 &
  \multicolumn{1}{c|}{0.8046} &
  \multicolumn{1}{c|}{0.8214} &
  \multicolumn{1}{c|}{0.7315} &
  \multicolumn{1}{c|}{0.8168} &
  \multicolumn{1}{c|}{0.8045} &
  0.7296 \\ \hline
{\color[HTML]{24292F} UPerNet} &
  \multicolumn{1}{c|}{0.8634} &
  \multicolumn{1}{c|}{0.7801} &
  \multicolumn{1}{c|}{0.7515} &
  \multicolumn{1}{c|}{0.9198} &
  \multicolumn{1}{c|}{0.773} &
  0.7704 &
  \multicolumn{1}{c|}{0.8414} &
  \multicolumn{1}{c|}{0.8043} &
  \multicolumn{1}{c|}{0.7569} &
  \multicolumn{1}{c|}{0.8473} &
  \multicolumn{1}{c|}{0.8103} &
  0.7575 \\ \hline
  {\color[HTML]{24292F} BiT} &
  \multicolumn{1}{c|}{0.8840} &
  \multicolumn{1}{c|}{\textbf{0.8647}} &
  \multicolumn{1}{c|}{\textbf{0.8741}} &
  \multicolumn{1}{c|}{-} &
  \multicolumn{1}{c|}{-} &
   - &
  \multicolumn{1}{c|}{0.8710} &
  \multicolumn{1}{c|}{\textbf{0.8758}} &
  \multicolumn{1}{c|}{\textbf{0.8734}} &
  \multicolumn{1}{c|}{-} &
  \multicolumn{1}{c|}{-} &
   - \\ \hline
\end{tabular}%
}
\end{table}

\section{EXPERIMENTS AND EVALUATION}
\label{sec:experiments}
Tasks like instance segmentation \cite{lin2014microsoft} and image segmentation \cite{lin2014microsoft} involves a similar pipeline like that of change detection. We used the change detection framework \cite{likyoocdp:2021} for conducting most of our CD-based experiments. The pytorch based change detection framework supports many models, encoders and deep architectures. In this change detection pipeline, the bi-temporal images are encoded into feature vectors using two encoders. The encoders can be either siamese or non-siamese type. The input images are encoded into encoded feature vectors, which are fused either by concatenating, summing, subtracting or finding the absolute difference of the vectors. The resulting feature vector are decoded as output which is compared to the ground truth mask. We apply simple transforms like flipping, scaling and cropping for augmentations of the dataset images at the global image level. Unlike other tasks, for change detection the same transform has to be applied to the multi-temporal images.

\subsection{Evaluation Criteria}
We report the precision, recall and F-score values as the metric to compare the performance in different change detection tasks under various encoder backbones settings which is defined as below:
\begin{equation}
\centering
Precision = \dfrac{TP} {TP + FP}
\end{equation}

\begin{equation}
\centering
Recall = \dfrac {TP} {TP + FN}
\end{equation}

\begin{equation}
\centering
F-Score = \dfrac {2\hspace{0.1cm}X\hspace{0.1cm}Precision\hspace{0.1cm}X\hspace{0.1cm}Recall} 
{Precision + Recall}
\end{equation}

where, TP is the number of true positives, FP is the number of false positives, TN is the number of true negatives and FN is the number of false negatives. We report the bounding box mean average precision (bbox mAP) 
and segmentation mean average precision (segm mAP) as the performance metric to compare the various object detection and instance segmentation methods under different backbone settings in the benchmark. Intersection Over Union (IOU) is a measure that evaluates the overlap between ground truth and predicted bounding boxes and helps to determine if a detection is True Positive.
\noindent
\textit{Average Precision} (AP) is obtained by interpolating the precision at each recall level $r$, taking the maximum precision whose recall value is greater than or equal to $r + 1$. Finally the mean of AP of all classes gives \textit{mean Average Precision} (mAP), the metric being used to compare different detectors.

\subsection{Experiments}
\begin{figure}
\centering
\begin{subfigure}{0.5\linewidth}
\centering
\includegraphics[width=.8\linewidth]{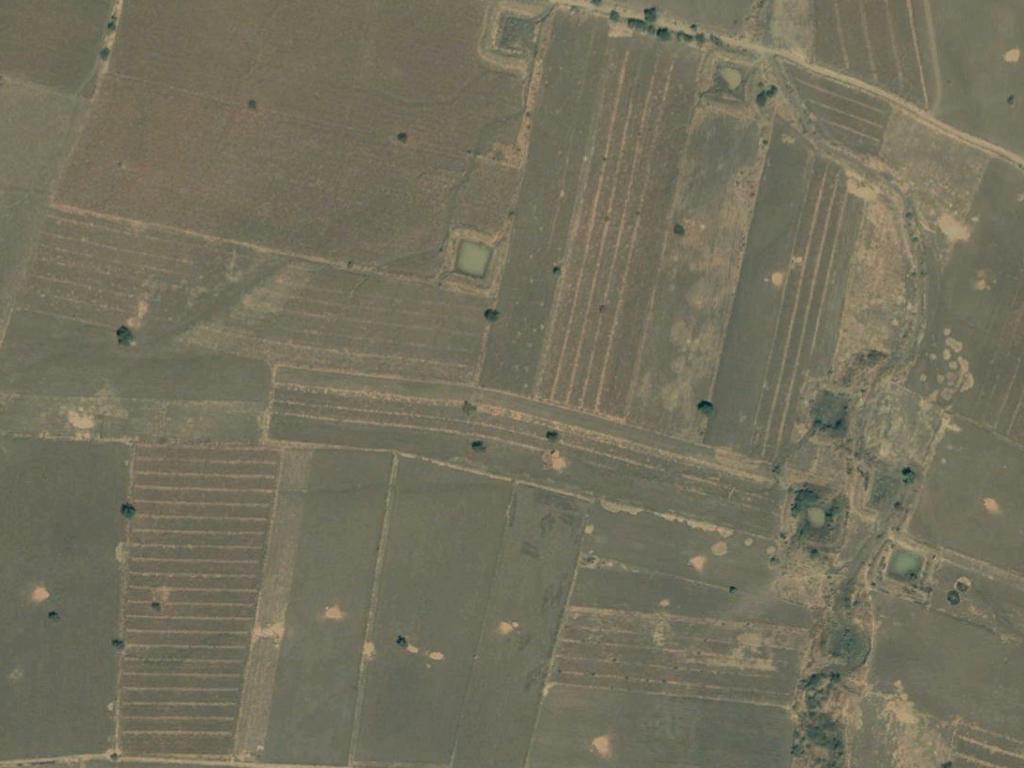}
\label{fig:sfig1}
\vspace{0.3cm}
\end{subfigure}%
\begin{subfigure}{.5\linewidth}
\centering
\includegraphics[width=.8\linewidth]{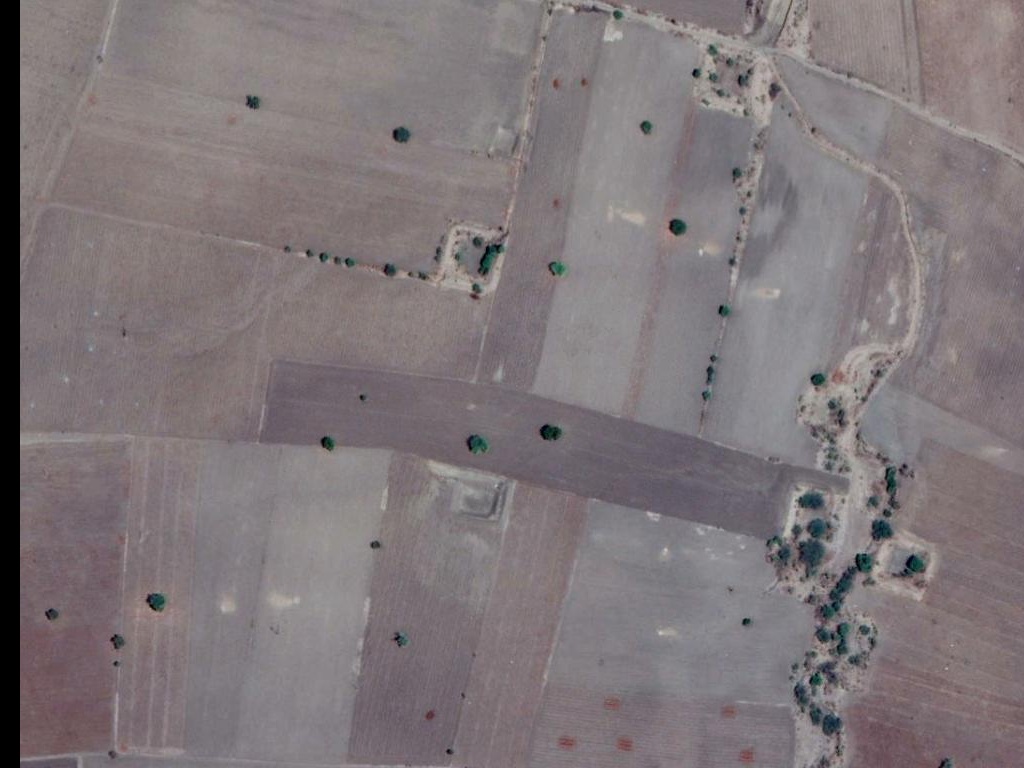}
\label{fig:sfig2}
\vspace{0.3cm}
\end{subfigure}
\begin{subfigure}{.3\linewidth}
\centering
\includegraphics[width=\linewidth]{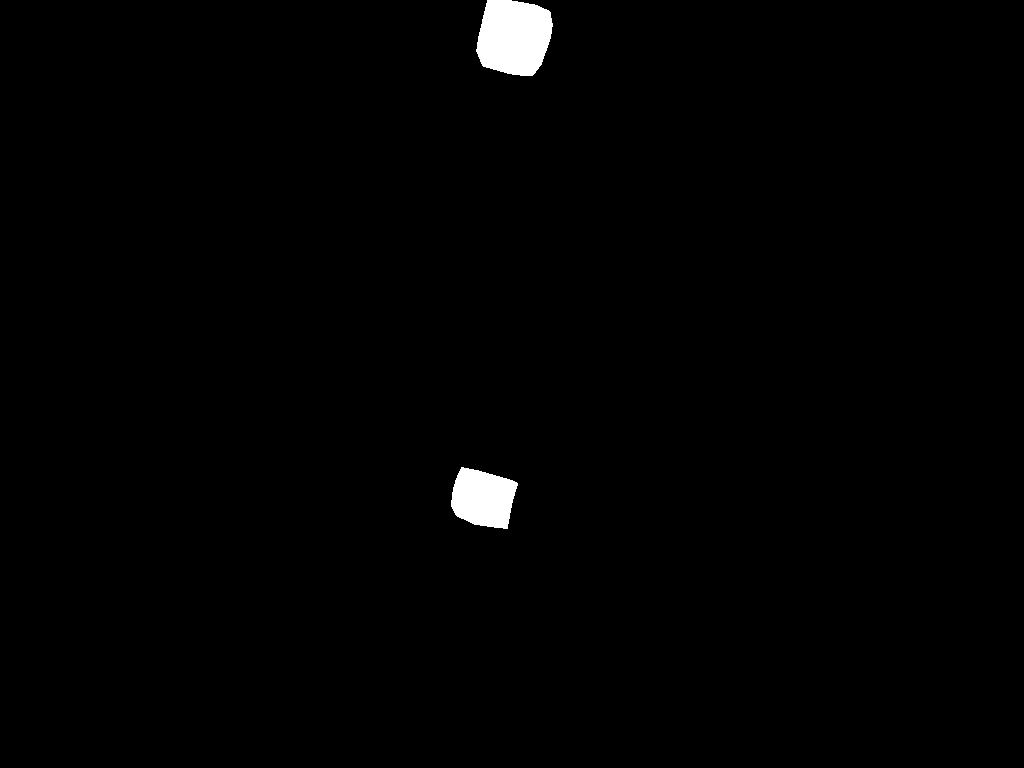}
\label{fig:sfig3}
\end{subfigure}
\begin{subfigure}{0.3\linewidth}
\centering
\includegraphics[width=\linewidth]{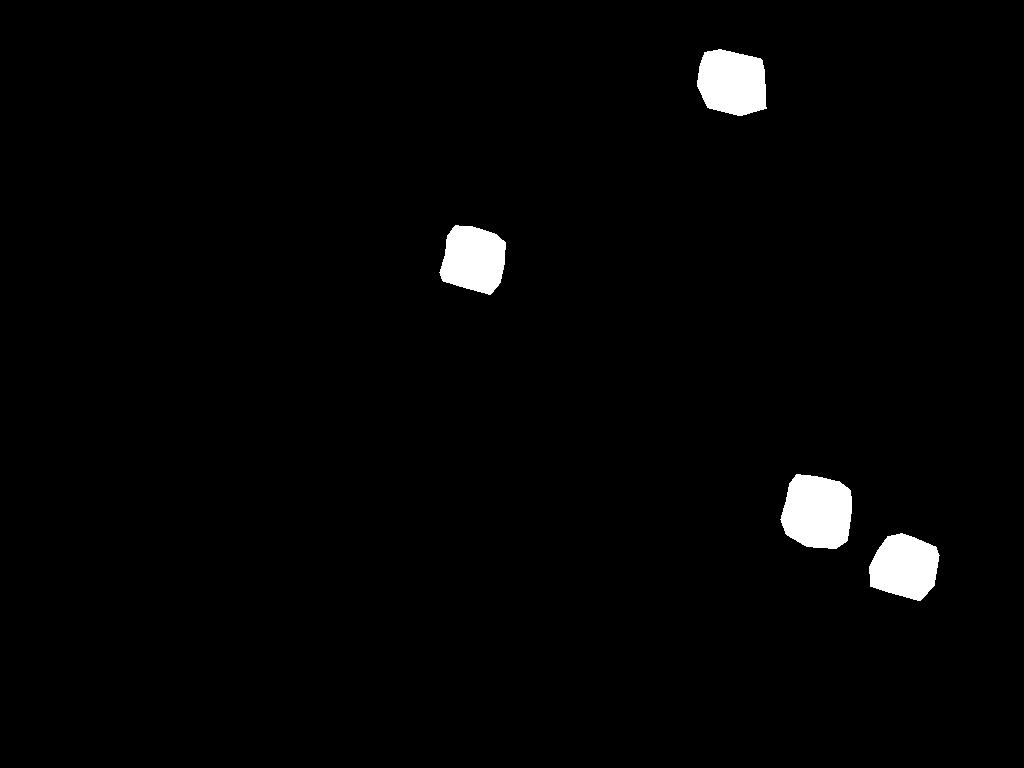}
\label{fig:sfig4}
\end{subfigure}
\begin{subfigure}{0.3\linewidth}
\centering
\includegraphics[width=\linewidth]{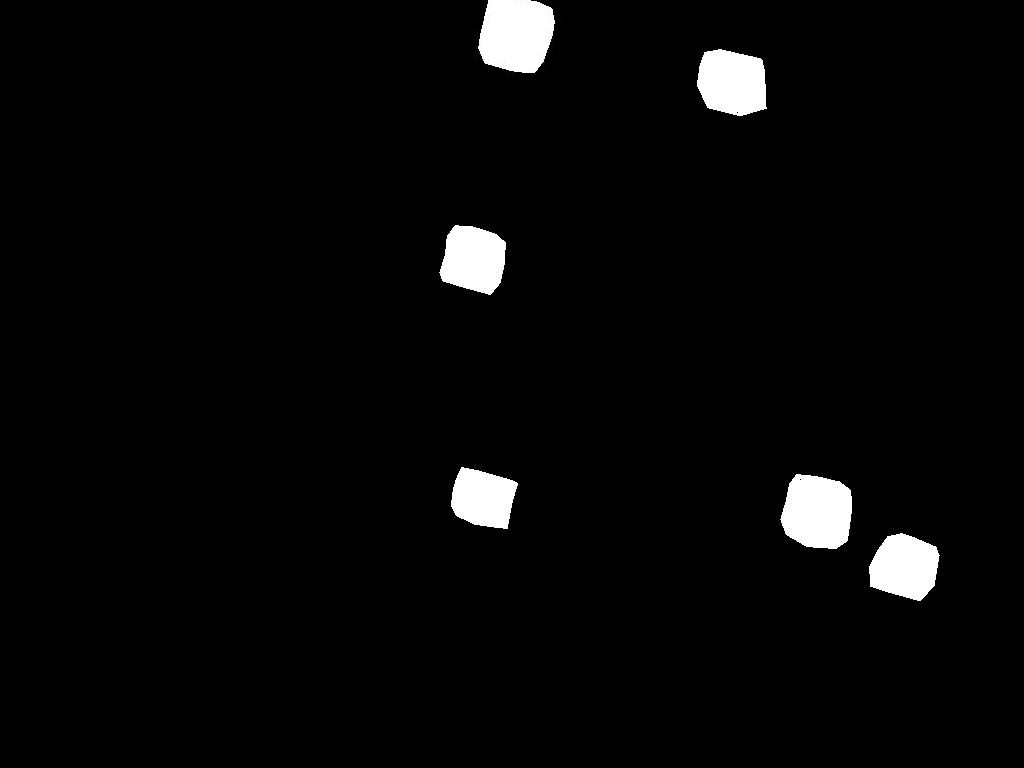}
\label{fig:sfig5}
\end{subfigure}
\caption{T0 and T1 input image pair in the top row and the image masks correspondingly for Task 1, Task 2 and Task 3 from left to right in the bottom row.}
\label{fig:task1_task2_task3}
\end{figure}

We benchmarked various object detection (OD) and instance segmentation (IS) methods on the farm pond OD/IS dataset. We use an existing encoder-decoder based change detection framework \cite{likyoocdp:2021} for the binary change detection tasks corresponding to Farm pond construction/ demolition (Task 1) and Farm pond dried/ wetted (Task 2). The encoder belongs to siamese type and the features from each branch are fused by concatenation. A few experiments were conducted to analyze the effect of various components of the change detection pipeline. We combined the masks from Task 1 and Task 2 to understand the impact of performance on multiple change classes. Combining the tasks lead to masks having more change objects than either of the tasks (Refer Fig \ref{fig:task1_task2_task3}). The train-validation split for each Task has been mentioned in Table \ref{tab:cd_train_test_split} . We split the farm ponds dataset for OD/IS task in a 80/20 train/test proportion. We used the MMDetection~\cite{mmdetection} framework for implementing the various methods in the OD/IS benchmark. We used the pretrained backbones of ResNet-50, ResNet-101 and MobileNetv2 primarily trained on Imagenet, COCO and cityscapes for the object detection experiments and Resnet-50, Resnet-101 and Swin Transformer pretrained backbones for instance segmentation experiments. 

\section{RESULTS AND OBSERVATIONS}
\label{sec:results}
The results of change detection on Farm pond construction/demolition (Task-1) taking ResNet-18, ResNet-50 and ResNet-101 as encoders are given in Table \ref{tab:results_task1}.
BiT model despite not having best precision or recall values performs the best in terms of F-score with the score of \textbf{0.8704}.
MANet despite performing highest in precision scores poorly in recall values leading to poor F-score and FPN scores the highest recall value.
FPN has the best recall in Resnet-101 and consistently better than its smaller counterparts, ResNet-18 and ResNet-50.
\par
The results of change detection on Farm pond dried/wetted (Task-2) taking ResNet-18, ResNet-50 and ResNet-101 as encoders are given in Table \ref{tab:results_task2}. 
We can notice that Unet has the  best F-Score of \textbf{0.9171} and the highest precision for ResNet-18.
For ResNet-50 encoder, we can observe that FPN has the best F-Score of \textbf{0.9125}, though DeepLabV3 has best recall and UperNet has best precision values.
DeepLabV3+ has the highest recall value with comparatively smaller precision than the best performing model. If we observe carefully, Task-2 has higher overall scores irrespective of the encoder architecture. This may be due to the presence of farm ponds at the same locations in both temporal images where the change is registered and also due to the fact that the number of positives are much smaller than the number of negative images for the task.

The results of combining change detection tasks (Task-1 and Task-2) as Task-3 taking ResNet-18, ResNet-50 and ResNet-101 as encoders  are given in Table \ref{tab: results_task3}. 
For the ResNet-18 encoder, we can notice that BiT has the best F-score of \textbf{0.8741} and the highest recall.
For the ResNet-50 encoder, we can notice that PSPNet has the best precision but with poor recall values affecting the F-score. 

We also conducted an ablation study on Task 3 to check if the absence of negative examples in the training set could affect the change detection task performed.
We can see that BiT achieves the best F-Score with ResNet-18 as the encoder, but we can also see that there is a decrease in precision when compared to the best model in the complete dataset, despite better recall values. 
The results for F-scores also show a similar trend, suggesting that even with a high ratio, the performance is comparable even in the absence of negative images in the train set. 
Similarly, we observe that UNet++ with the best F-Score trails the model trained on the entire dataset by only a few points; this further supports the idea that positive samples aid in learning while negative images may increase the model's robustness. 
\par
In this section, we analyze the performance of object detection and instance segmentation task on the Farm pond OD/IS dataset. The results of comparison of object detection on various existing models are given in Table \ref{tab:results_od}. 
For the object detection task, Probabilistic Anchor Assignment (PAA) model achieves the best performance of \textbf{0.575} mAP with IoU(0.5:0.95) with the ResNet-50 backbone. We can notice that empirical attention with the attention component (0010) achieves the best mAP under IoU(0.75). Guided Anchoring with FasterRCNN model performs the best in bbox mAP(0.5) using ResNet-50 backbone. For the instance segmentation task, many models have the capability to generate segmentation masks along with bounding boxes results and the COCO metric for both are given in the Table \ref{tab:results_is}. We can notice that CascadeRCNN with ResNet-50 performs the best with \textbf{0.577} bbox mAP(0.5:0.95) and \textbf{0.694} bbox mAP(0.75) indicating it to be a robust detector. Swin Transformer has the best bbox mAP(0.5) with its own independent backbone. For instance segmentation, Deformable ConvNetsv2 with MRCNN model using ResNet-50 as backbone performs the best on segm mAP metric with IoU(0.5:0.95), while MaskRCNN with ResNet-50 performs the best in segm mAP(0.75) and Swin Transformer performs the best under the segm mAP(0.5) metric.

\begin{table}[h]
\centering
\caption{Instance segmentation results on OD/IS dataset.}
\label{tab:results_is}
\resizebox{\columnwidth}{!}{%
\begin{tabular}{|c|c|c|c|c|c|c|c|}
\hline
\textbf{Method} &
  \textbf{Backbone} &
  \textbf{\begin{tabular}[c]{@{}c@{}}bbox mAP \\ (0.5:0.95)\end{tabular}} &
  \textbf{\begin{tabular}[c]{@{}c@{}}bbox mAP \\ (0.5)\end{tabular}} &
  \textbf{\begin{tabular}[c]{@{}c@{}}bbox mAP \\ (0.75)\end{tabular}} &
  \textbf{\begin{tabular}[c]{@{}c@{}}segm mAP \\ (0.50:0.95)\end{tabular}} &
  \textbf{\begin{tabular}[c]{@{}c@{}}segm mAP \\ (0.5)\end{tabular}} &
  \textbf{\begin{tabular}[c]{@{}c@{}}segm mAP \\ (0.75)\end{tabular}} \\ \hline
\begin{tabular}[c]{@{}c@{}}Mask RCNN \end{tabular} &
  ResNet - 50 &
  0.558 &
  0.834 &
  0.68 &
  0.59 &
  0.853 &
  \textbf{0.696} \\ \hline
\begin{tabular}[c]{@{}c@{}}Swin Transformer \end{tabular} &
  \begin{tabular}[c]{@{}c@{}}Swin  Transformer\end{tabular} &
  0.563 &
  \textbf{0.851} &
  0.667 &
  0.59 &
  \textbf{0.874} &
  0.671 \\ \hline
\multirow{2}{*}{\begin{tabular}[c]{@{}c@{}}Cascade RCNN \end{tabular}} &
  ResNet - 50 &
  \textbf{0.577} &
  0.81 &
  0.688 &
  0.575 &
  0.829 &
  0.693 \\ \cline{2-8} 
 &
  ResNet - 101 &
  0.563 &
  0.813 &
  \textbf{0.694} &
  0.572 &
  0.827 &
  0.685 \\ \hline
\begin{tabular}[c]{@{}c@{}}Hybrid Task Cascade \end{tabular} &
  ResNet - 50 &
  0.533 &
  0.785 &
  0.663 &
  0.541 &
  0.805 &
  0.643 \\ \hline
\begin{tabular}[c]{@{}c@{}}Deformable \\ ConvNets v2 (MRCNN) \end{tabular} &
  ResNet - 50 &
  0.571 &
  0.833 &
  0.673 &
  \textbf{0.591} &
  0.859 &
  0.691 \\ \hline
\end{tabular}%
}
\end{table}

\section{CONCLUSION}
\label{sec:conclusion}
The availability of high-resolution images due to advances in remote sensing capabilities has led to the use of deep learning models for change detection. 
In this paper, we introduced \textbf{FPCD}, 
a publicly available dataset for change detection tasks for minor irrigation structures - farm ponds. 
We also introduced a small-scale public dataset for object detection and instance segmentation of four farm pond categories. Future works can address the issues associated with limited data availability and class imbalance issues in the dataset.
\begin{table}[h]
\centering
\caption{Object detection results on OD/IS dataset.}
\label{tab:results_od}
\resizebox{\columnwidth}{!}{%
\begin{tabular}{|c|c|c|c|c|c|}
\hline
\textbf{Method} &
  \textbf{Model Name} &
  \textbf{Backbone} &
  \textbf{\begin{tabular}[c]{@{}c@{}}bbox mAP\\ (0.5:0.95)\end{tabular}} &
  \textbf{\begin{tabular}[c]{@{}c@{}}bbox mAP \\ (0.5)\end{tabular}} &
  \textbf{\begin{tabular}[c]{@{}c@{}}bbox mAP\\ (0.75)\end{tabular}} \\ \hline

\begin{tabular}[c]{@{}c@{}}Grid RCNN \end{tabular} &    
  \begin{tabular}[c]{@{}c@{}}Grid RCNN\end{tabular} &
  ResNet - 50 &
  0.56 &
  0.553 &
  0.551 \\ \hline  

\begin{tabular}[c]{@{}c@{}}Deformable DETR \end{tabular} &
\begin{tabular}[c]{@{}c@{}} 2-stg Def. DETR\end{tabular} &
  ResNet - 50 &
  0.526 &
  0.774 &
  0.624 \\ \hline
\begin{tabular}[c]{@{}c@{}}Faster RCNN\end{tabular} &
  \begin{tabular}[c]{@{}c@{}}Faster RCNN\end{tabular} &
  ResNet - 50 &
  0.555 &
  0.832 &
  0.646 \\ \hline
\multirow{2}{*}{\begin{tabular}[c]{@{}c@{}}Empirical Attention \end{tabular}} &
  \begin{tabular}[c]{@{}c@{}}AC-0010\end{tabular} &
  ResNet - 50 &
  0.553 &
  0.838 &
  \textbf{0.687} \\ \cline{2-6} 
 &
  \begin{tabular}[c]{@{}c@{}}AC-1111\end{tabular} &
  ResNet - 50 &
  0.557 &
  0.829 &
  0.663 \\ \hline
  \begin{tabular}[c]{@{}c@{}}GFL \end{tabular} &
  GFL &
  ResNet - 50 &
  0.553 &
  0.815 &
  0.653 \\ \hline
  \begin{tabular}[c]{@{}c@{}}GH SSD \end{tabular} &
  GHM &
  ResNet - 50 &
  0.528 &
  0.787 &
  0.629 \\ \hline
\multirow{2}{*}{\begin{tabular}[c]{@{}c@{}}Guided Anchoring \end{tabular}} &
  \begin{tabular}[c]{@{}c@{}}Faster RCNN\end{tabular} &
  ResNet - 50 &
  0.566 &
  \textbf{0.843} &
  0.672 \\ \cline{2-6} 
 &
  RetinaNet &
  ResNet - 50 &
  0.547 &
  0.832 &
  0.637 \\ \hline
  \begin{tabular}[c]{@{}c@{}}PAA \end{tabular} &
  PAA &
  ResNet - 50 &
  \textbf{0.575} &
  0.832 &
  0.686 \\ \hline
\begin{tabular}[c]{@{}c@{}}YOLOv3 \end{tabular} &
  YOLOv3 &
  MobileNetv2 &
  0.437 &
  0.729 &
  0.485 \\ \hline
\end{tabular}%
}
\end{table}

\bibliographystyle{apalike}
{\small
\bibliography{example}}
\end{document}